%% file: main.tex
\documentclass[10pt,twocolumn,letterpaper]{article}

\usepackage{cvpr}              

\usepackage{graphicx}
\usepackage{amsmath}
\usepackage{amssymb}
\usepackage{booktabs}

\usepackage{fullpage}
\usepackage{multicol}
\usepackage{enumitem}
\usepackage{cuted}

\usepackage[pagebackref,breaklinks,colorlinks]{hyperref}

\usepackage[capitalize]{cleveref}
\crefname{section}{Sec.}{Secs.}
\Crefname{section}{Section}{Sections}
\Crefname{table}{Table}{Tables}
\crefname{table}{Tab.}{Tabs.}

\def\cvprPaperID{11} 
\def\confName{CVPR}
\def\confYear{2023}

\begin{document}


\title{LPMM : Intuitive Pose Control for Neural Talking-Head Model via Landmark-Parameter Morphable Model}

\author{
Kwangho Lee\thanks{Equal contribution} \qquad
Patrick Kwon\footnotemark[1] \qquad
Myung Ki Lee \qquad
Namhyuk Ahn \qquad
Junsoo Lee\\
Naver Webtoon AI\\
{\tt\small \{superior.kwangho.lee, patrick.kwon, myunggi, nhahn, junsoolee93\}@webtoonscorp.com}
}
\maketitle

\begin{strip}
\vspace*{-1.5cm}
\centering
 \includegraphics[width=\textwidth]{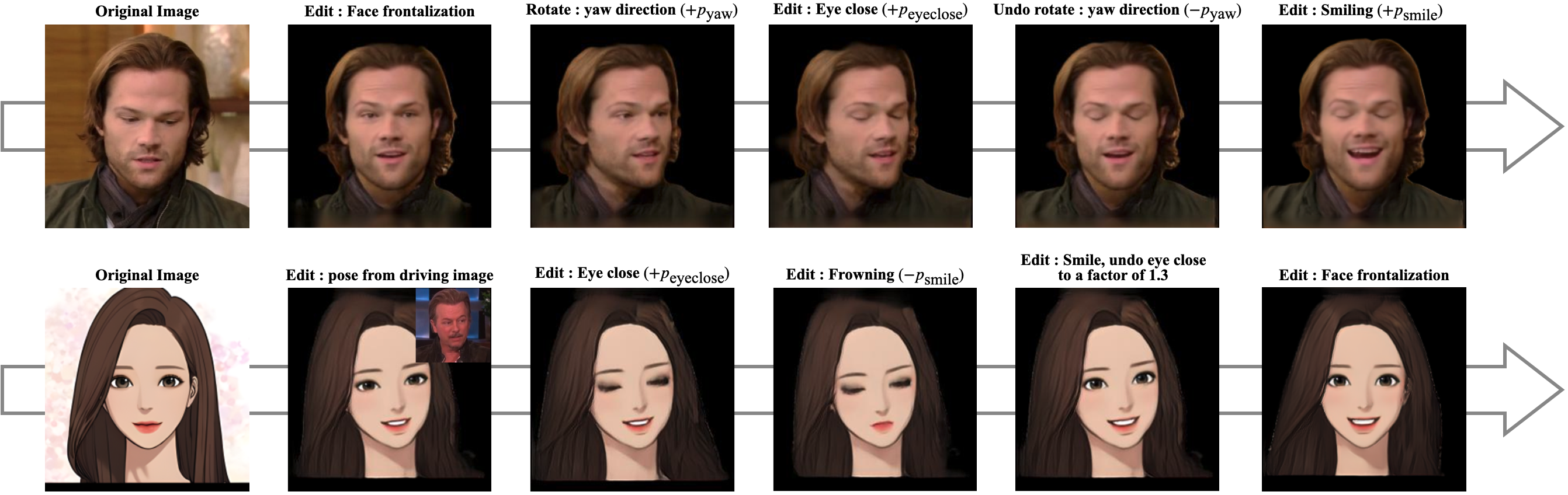}
\captionof{figure}{LPMM allows user-friendly pose control over portrait images, by translating facial landmarks to the parametric domain. This enables a sequential, intuitive editing of facial expressions and head orientation, either without a driving source image (top row) or with a driving source image (bottom row).}
\label{fig:teaser}
\end{strip}

\input{section/0_abstract}

\input{section/1_intro}

\input{section/2_related}

\input{section/3_method}
\input{section/4_experiment}
\input{section/5_conclusion}

{\small
\bibliographystyle{ieee_fullname}
\bibliography{reference}
}

\end{document}

%% file: section/0_abstract.tex
\begin{abstract}
\vspace{-1em}
\vfill
While current talking head models are capable of generating photorealistic talking head videos, they provide limited pose controllability. 
Most methods require specific video sequences that should exactly contain the head pose desired, being far from user-friendly pose control. 
Three-dimensional morphable models (3DMM) offer semantic pose control, but they fail to capture certain expressions.
We present a novel method that utilizes parametric control of head orientation and facial expression over a pre-trained neural-talking head model.
To enable this, we introduce a landmark-parameter morphable model (LPMM), which offers control over the facial landmark domain through a set of semantic parameters. 
Using LPMM, it is possible to adjust specific head pose factors, without distorting other facial attributes. 
The results show our approach provides intuitive rig-like control over neural talking head models, allowing both parameter and image-based inputs.
\end{abstract}


%% file: section/1_intro.tex
\section{Introduction}
\label{sec:intro}


Photorealistic facial reenactment research has achieved remarkable improvements in both perceptual quality and frame consistency. 
Current deep-learning based facial reenactment methods ~\cite{burkov2020@lpd, wang2021@lia, wang2021@osfv, zakharov2019@fsal} can synthesize high-quality videos of a talking face, using a target human identity and animation sources. 
Such systems hold great potential for various applications such as telepresence and virtual avatar generation.


Despite these advancements, current talking head models have limitations in controlling facial content.
Most methods require target video sequences that should exactly contain the desired head pose\footnote{Following Burkov \textit{et al}.~\cite{burkov2020@lpd}, for this paper, we use the notation `head pose' to refer to both head orientation and facial expression.}, usually represented in the form of facial landmarks~\cite{zakharov2019@fsal, Suwajanakorn2017SynthesizingO, Kim2018DeepVP, Wang2018VideotoVideoS}. 
One intuitive method of pose control is to let the user upload their facial videos to the system, but for real-world services, many users felt uncomfortable sharing their own videos and preferred to use existing video sources. 
This, in turn, places an additional burden on the user, as they must find and provide such video sources.


To enable better control over the facial domain, several methods focused on providing facial synthesis in the form of three-dimensional, user-friendly control (also known as \textit{face rig)}. 
Those models focused on using a three-dimensional morphable face model (3DMM)~\cite{blanz1999@3dmm}, which allows the user to control over various facial semantic parameters, such as identity, expressions, texture, illuminance, and head orientation. 
However, most 3DMM-based researches are focused on generating 3D animations instead of photorealistic videos. 
One main reason is that 3DMMs tend to be bound by the lack of facial training data, which requires complex 3D facial scanning, resulting in a lack of photorealism.

To be best of our knowledge, StyleRig ~\cite{tewari2020@stylerig} is the first model to provide rig-like control over photorealistic portrait images, by adding 3DMM's semantic controllability over StyleGAN ~\cite{karras2019stylegan}, a GAN-based image generator. 
Requiring only a pre-trained StyleGAN model, StyleRig achieved an intuitive, rig-like control over high quality portrait images. 
However, StyleRig has limited expressiveness, as it fails to produce certain head poses such as in-plane head rotation and asymmetric expressions. 
Moreover, as StyleGAN does not explicitly disentangles identity and pose information, \textit{identity bleeding} can occur during high-level expression editing, i.e. the resulting face's identity changes when it should not.


Building upon such limitations, we propose a novel solution for parameter-based neural talking head synthesis, by combining the advantages of both neural talking head methods and parameter-based pose control into a single method. 
We focus on expressing head poses as semantic parameters, which are transformed into latent codes for the given talking head generator to create photorealistic facial images with or without additional driving images. 
We use a fixed, pretrained talking head model, and do not require additional data for training.

Our main contribution is the landmark-parameter morphable model (LPMM), a model designed to connect landmarks to the parametric domain in which the user can adjust different facial expressions and head pose in a meaningful manner. 
Since LPMM is built upon a large and diverse facial landmark dataset, which can be easily collected compared to facial scans, it achieves a better generalization over expression diversity and semantic parameterization compared to 3DMM. 
The usage of a talking head generator instead of a 3d renderer ensures photorealistic results of our method. 
The results display that our approach successfully provides intuitive rig-like control for a pre-trained neural talking head model, while still allowing traditional facial image inputs. 

In summary, we present the following contributions:
\vspace{-2mm}
\begin{itemize}[itemsep=0pt]
    \item Novel pipeline that can give pose controllability to talking head model, which does not require additional training for fixed model.
    \item A method to provide additional rig-like control while maintaining the inference method of the existing talking head model. Therefore, the performance of the generator can be fully utilized.
    \item We show that our pipeline is independent of model architecture and can be applied to arbitrary latent-based talking head generators.
\end{itemize}

%% file: section/2_related.tex
\section{Related Work}
\label{sec:related}

 \noindent\textbf{Neural Talking Head} Most models focus on extracting pose information from a driving video source into the form of latent codes and applying it to a facial identity ~\cite{Nirkin2019FSGANSA, Siarohin2019FirstOM, Suwajanakorn2017SynthesizingO}. Doing so minimizes the identity-pose disentanglement issue, making it suitable for further pose editing. Such methods, however, do not provide additional means for the user to control head pose and facial expression. Some talking head methods accept user-specified head rotation input ~\cite{wang2021@osfv}, but do not allow the user to adjust facial expressions in a semantic manner. Other methods rely on audio inputs to control the lower part of the face (lips, jaw movement) ~\cite{Yi2020AudiodrivenTF, KR2020ALS, Chen2019HierarchicalCT} but fail to control other parts of the head.

\noindent\textbf{3D Morphable Model} To allow user-friendly control of facial contents, many recent works utilize 3D Morphable Models (3DMM)~\cite{blanz1999@3dmm, Paysan2009A3F, li2017flame}. 3DMMs parameterize faces into different factors (identity, expressions, texture, illumination, pose), which allows for the rendering of plausible human faces of varying identity and expression. 
Creating such a model, however, requires a substantial amount of 3D facial scans, which tend to lack photo-realism in the final rendering, and have limited generalization over semantic parameters~\cite{tewari2020@stylerig}. While most 3DMM-based literature has focused on generating 3d animations, several neural-talking head methods \cite{Kim2018DeepVP, Yi2020AudiodrivenTF} use 3DMM as a facial basis, but they are limited in output resolution and degrades in visual quality when reenacting extreme expressions.

\begin{figure*}[t!]
  \centering
  \includegraphics[width=1\linewidth]{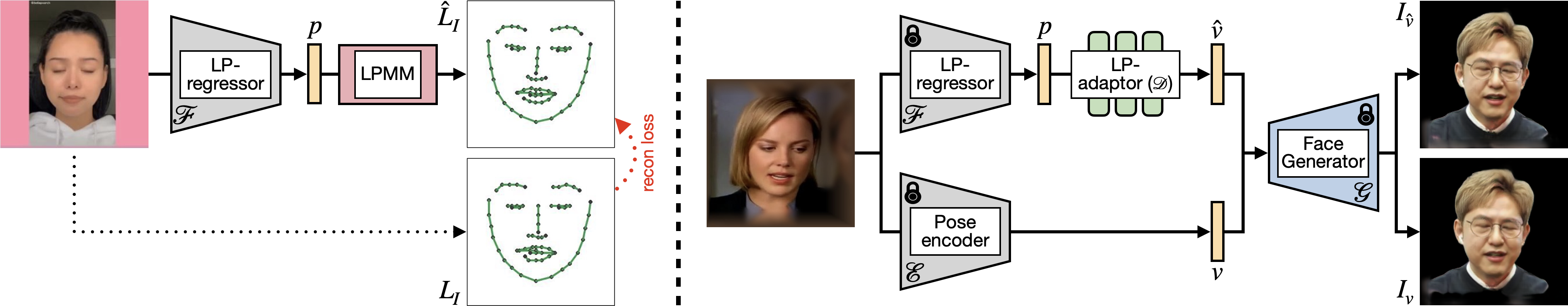}
   \caption{The training of our model is divided into two stages. (Left) The LP-regressor processes the input facial image to generate LPMM parameters, and is trained so that the reconstructed facial landmark matches the original. (Right) The LP-adaptor is used to transform LPMM parameters into the latent space of a pretrained talking head model's pose encoder. While training LP-adaptor, all weights other than LP-adaptor itself are frozen.}
   \label{fig:Overview}
\end{figure*}

\noindent\textbf{StyleGAN-based Model} Tewari \textit{et al.} ~\cite{tewari2020@stylerig} used StyleRig, a new approach to merge 3DMM's parametric control with StyleGAN~\cite{karras2019stylegan, karras2020stylegan2}, a model is known for photorealistic portrait image generation. While several methods focused on editing StyleGAN-generated images by disentangling different attributes from the latent space of StyleGAN~\cite{richardson2021psp, roich2022pti, tov2021e4e, alaluf2021restyle}, most methods failed on disentangling human identity from a pose. Instead, StyleRig is based on a self-supervised cycle consistency loss and a differentiable face renderer, leading to high-quality facial images while maintaining control over expression, illumination, and pose. However, StyleRig fails to exploit 3DMM's full expression space, resulting in incorrect expression mappings for the final result (e.g. in-place head rotation, eye-blinking). Its visual quality is also limited by the face renderer used. Our work is highly motivated by StyleRig's approach of adopting parametric control over an image generator, but differs in the model architecture, semantic controllability, and a novel morphable model.




%% file: section/3_method.tex
\section{Method}
\label{sec:method}


Our solution achieves an explicit rig-like control over the head pose of facial images generated from neural talking-head models. Based on a talking head generator ~\cite{burkov2020@lpd, wang2021@lia} that generates a facial image $I_w$ using latent code $w$, our approach focuses on providing additional pose editing through the usage of semantic parameters, while still allowing traditional pose image inputs. Prior to training, we prepare a landmark-parameter morphable model (LPMM), which is designed to control head pose information based on a set of control parameters (Sec. \hyperref[subsec:LPMM]{3.1}). Those parameters are linked to significant components in the facial landmark domain, calculated through a standard PCA decomposition ~\cite{jolliffe2002pca}.

Given a facial image $I$, a landmark-parameter regressor (LP-regressor) is used to acquire the respective parameter values for LPMM, which are then edited to convey the desired head pose (Sec. \hyperref[subsec:LP-regressor]{3.2}). The modified parameters are then sent to the landmark-parameter adaptor (LP-adaptor), and converted to a latent code for the selected talking head generator (Sec. \hyperref[subsec:LP-adaptor]{3.3}). One practical use case would be to edit a face such that its eyes are closed, without altering other elements of the face. In such a case, the user can simply adjust the parameter values corresponding to the eyes, without using additional driving image inputs.

\begin{figure}[t]
  \centering
  \includegraphics[width=1.0\linewidth]{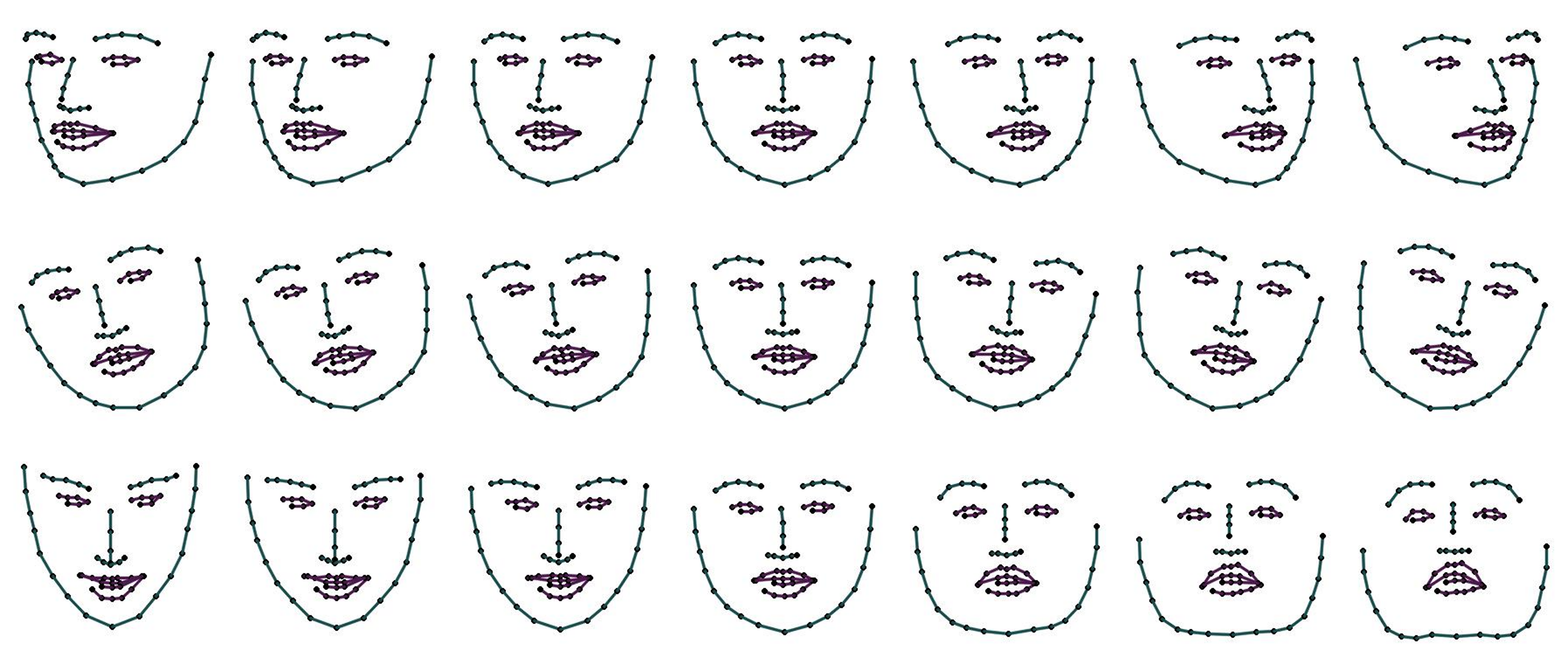}
   \caption{Interpolation parameters of LPMM. Each row shows that different parameter can control head pose independently. The average face landmark $\bar{L}$ in the middle column represents a expression-neutral, frontalized face.}
   \label{fig:lpmm_interpolation}
\end{figure}

\subsection{Landmark-Parameter Morphable Model}
\label{subsec:LPMM}

Our proposed model attempts parametric control on the facial landmark, which serves as a good basis of head pose for many talking head models~\cite{zakharov2019@fsal, Suwajanakorn2017SynthesizingO, Kim2018DeepVP, Wang2018VideotoVideoS}.
Since facial landmarks $L=(x_{1},y_{1},x_{2},y_{2},.....,x_{n},y_{n})\in\mathbb{R}^{2n}$ consist of $n=68$ points that connect to different facial components (such as eyes, nose, mouth), we assume that through a linear combination of exemplar facial landmark components, it is possible to generate an arbitrary head pose, leading towards an abstract pose-specific representation. 
Moreover, facial landmarks can be acquired from a wider range of 2D facial video data, unlike 3D facial scans, so we believe LPMM will result in a highly generalized model compared to 3DMM.

To extract such exemplar components, we perform Principal Component Analysis (PCA)~\cite{jolliffe2002pca} upon the facial landmark data~\cite{Nagrani17@vox1, bulat2017@face_landmark_detector}. Being a common technique for data compression, PCA performs an eigen-decomposition operation over the data covariance matrix, extracting the most significant eigenvectors in the form of linearly independent components. Since the first principal components created from PCA explain the most variance of the data compared to the latter components, we assumed that these first components will be responsible for the head orientation (yaw, pitch, roll) movements, which tend to show the high magnitude of movement in the coordinate system while being orthogonal to one another. Other facial expressions, such as eye-blinking, have movements of lower magnitude, thus were assumed to be linked with the latter components. 

We collected a large number of facial landmarks from talking-head video datasets~\cite{Nagrani17@vox1, chung2018voxceleb2}. Since LPMM should not include components that are linked to head translation movements, each video frame was preprocessed to ensure that the face was aligned in the center of a square frame. Through PCA, $m$ principal components were calculated from the facial landmark dataset.

We define the morphable model as the set of facial landmarks, parameterized by the coefficients $p$. New arbitrary face landmarks can now be generated by the summation of average face landmark $\bar{L}$ and the linear combination of the parameters $p$ and eigenvectors $e\in\mathbb{R}^{2n}$. In this case, the maximum number of parameter $m$ is $2n$. 
\begin{equation}\label{eq:LPM}
    L_{new} = \bar{L} + \sum_{i=1}^{k}p_{i}e_{i}.
\end{equation}

After setting up the model, we discovered that the beginning components were indeed linked to head orientation movements, and the latter components were associated with other facial expressions, thus proving the initial assumption right (Figure \ref{fig:lpmm_interpolation}). Visualization of the different PCA components can be found in Supp. Mat.


\subsection{LP-regressor}
\label{subsec:LP-regressor}

As shown in Figure ~\ref{fig:Overview} (left), the landmark-parameter regressor (LP-regressor) can be seen as a function that estimates LPMM parameters from a given facial image. While LPMM is capable of generating arbitrary facial landmarks, it is required to estimate specific parameters to reconstruct a target landmark that successfully conveys the correct head pose.

Given an image $I\in\mathbb{R}^{H\times W\times3}$, LP-regressor $\mathcal{F}:\mathbb{R}^{H\times W\times3}\rightarrow\mathbb{R}^{k}$ calculates the corresponding LPMM parameters $p=[p_{1},p_{2},.....,p_{k}]$ up to a specific degree $k$ (maximum of $m$), which are used to reconstruct the image’s facial landmark $\hat{L}_{I}$. Since these parameters have different levels of significance in modeling facial landmarks, $k$ functions as a regulation term between expressiveness and complexity. We model LP-regressor based on the MobilenetV2 architecture~\cite{Sandler2018@mobilenetv2}, and trained it using $\ell_1$-loss between the original facial landmark ${L}_{I}$ and the reconstructed landmark $\hat{L}_{I}$.

This model can be viewed as a modification of previous facial landmark detectors, but instead of detecting coordinates of face landmarks, ours focus on imprinting head pose information into the LPMM parameters for intuitive pose control. Also, facial landmark data are not used as input for inference, and only their parameter counterparts function as the input for our system.

\begin{figure}[t]
  \centering
  \includegraphics[width=1\linewidth]{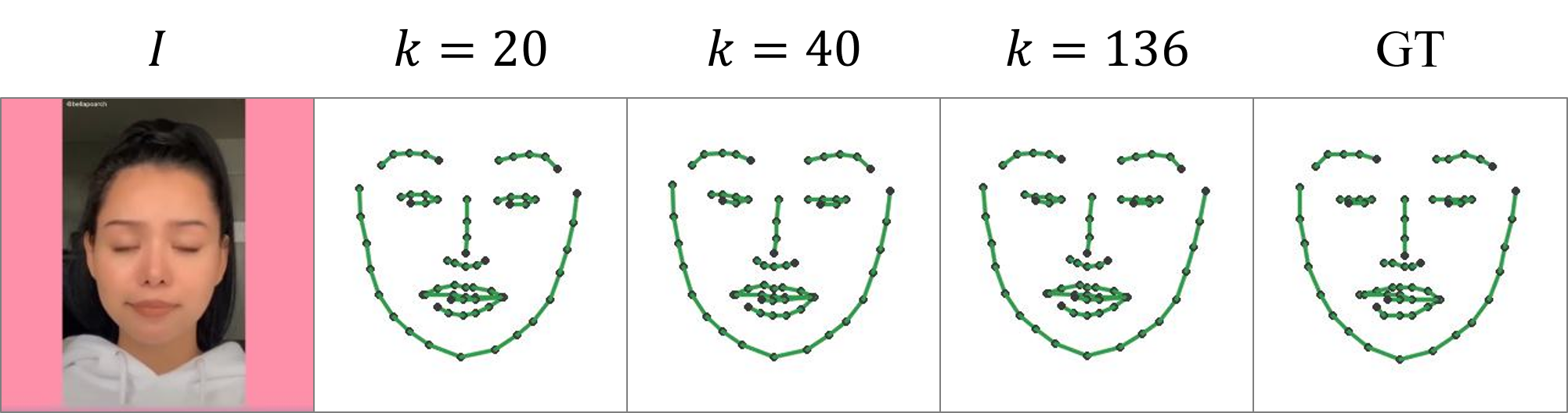}
   \caption{For LP-regressor, it is noted that the expressiveness of facial landmarks reconstructed from parameters change with differing values of $k$. The reconstructed landmarks cannot express eye-closing until $k=40$, and the accuracy difference between $k=40$ and $k=136$ is minor.}
   \label{fig:LP-regressor_results}
\end{figure}

From experimentation, we found the performance at $k=40$ is suitable for arbitrary facial landmark representation, and chose it as our default setting (Figure \ref{fig:LP-regressor_results}). For further implementation details, please refer to the experiments section. (Sec. \ref{sec:experiment})

\subsection{LP-adaptor}
\label{subsec:LP-adaptor}


Given a set of LPMM parameters, the landmark-parameter adaptor (LP-adaptor) generates latent codes for the pretrained talking head generator. Its main objective is to ensure that the parametric control over the facial landmark domain is also maintained for the generated face images from the neural talking head model. Previous latent-based facial manipulation methods ~\cite{richardson2021psp, tov2021e4e, alaluf2021restyle, roich2022pti} require finding new editing directions within the latent space, which is time-consuming and might lead to undesired pose distortion. On the other hand, LP-adaptor is capable of editing selected head pose, without distorting other facial attributes. 

We denote pose encoder of the talking head model~\cite{burkov2020@lpd,wang2021@lia} as $\mathcal{E}:\mathbb{R}^{H\times W\times3}\rightarrow\mathbb{R}^{w}$, which outputs an identity-agnostic pose vector $v\in\mathbb{R}^{w}$ for image $I$. And we denote its generator as $\mathcal{G}:\mathbb{R}^{w}\rightarrow\mathbb{R}^{H\times W\times3}$, which generates facial images with head pose from $I$. Figure \ref{fig:Overview}, right shows the overall framework of LP-adaptor. Following StyleRig ~\cite{tewari2020@stylerig}, we model LP-adaptor (denoted as $\mathcal{D}:\mathbb{R}^{k}\rightarrow\mathbb{R}^{w}$) as a three-layer linear perceptron (MLP) with ELU activations for every intermediate layer. 

Given an estimated parameter $p$ from LP-regressor, the last layer of MLP outputs $d$, which is added to the pre-calculated average pose vector $\bar{v}$ to form the final estimated pose vector $\hat{v} = \bar{v} + d$. The objective of LP-adaptor is to encode $\hat{v}$ such that it’s mapping in the latent space of pose encoder $\mathcal{E}(I)$ is in the right location; where the original pose vector $v$ is located.

We want the identity-specific features of output images to be consistent when controlling head pose using landmark parameters. Previous works~\cite{burkov2020@lpd,zakharov2019@fsal} discussed that using facial landmarks on neural talking head tasks may induce identity-bleeding issues. However, in our settings, facial landmarks themselves are not used as input. Also, we used a fixed identity embedding vector during the training pipeline, so that the LP-adaptor can only focus on the pose latent space without requiring any complex methods for the identity-pose disentanglement issue (i.e cycle-consistent per-pixel editing loss~\cite{tewari2020@stylerig}). And since the controllability is not limited towards a specific identity, it is not required for the LP-adaptor to be re-trained for a different human identity.




\noindent\textbf{Training LP-adaptor.} Figure \ref{fig:Overview} right shows the overall training of LP-adaptor. The training loss consists of a pixel-wise RGB loss $\mathcal{L}_{\text{rgb}}$ and a pose regularization loss $\mathcal{L}_{\text{pose-reg}}$ where $\lambda_{\text{rgb}}$ and $\lambda_{\text{pose-reg}}$ are fixed weights for each losses. When training LP-adaptor, all other networks except LP-adaptor (i.e. $\mathcal{F,E,G}$) are fixed.
\begin{equation}\label{eq:loss_total}
    \mathcal{L}_{total} = \lambda_{\text{rgb}}\mathcal{L}_{\text{rgb}} + \lambda_{\text{pose-reg}}\mathcal{L}_{\text{pose-reg}}.
\end{equation}

\noindent\textbf{Pixel-Wise RGB loss.} $\mathcal{L}_{rgb}$ is the $\ell_1$-loss between the generated image ${I}_{\hat{v}}=\mathcal{G}(\hat{v})$ using $\hat{v}$ from LP-adaptor, and the generated image ${I}_{{v}}=\mathcal{G}(v)$ using $v$ from the original pose encoder. While we could design the regression loss directly in latent space, this has been shown to not be very effective ~\cite{tewari2017mofa, tewari2020@stylerig}. 
\begin{equation}\label{eq:loss_rgb}
    \mathcal{L}_{rgb} = \|\mathcal{G}(\mathcal{D}(p^{(I)})) - \mathcal{G}(\mathcal{E}(I))\|_1.
\end{equation}

\noindent\textbf{Pose Regularization loss.} In order to ensure the existence of a ``base pose", we add an additional regularization loss. We enforce residual value $d$ in  $\hat{v} = \bar{v} + d$ should be zero when LP-adaptor got a parameters of average pose image $\mathcal{F}(\mathcal{G}(\bar{v}))$. Since the weights of $\mathcal{F}$ and $\mathcal{E}$ are frozen, this constraint enforces the mapping between the parametric and latent spaces.
\begin{equation}\label{eq:loss_posereg}
    \mathcal{L}_{\text{pose-reg}} = \|\mathcal{D}(\mathcal{F}(\mathcal{G}(\bar{v})))\|_1.
\end{equation}





\begin{figure*}[t!]
    \centering
    \includegraphics[width=1.0\linewidth]{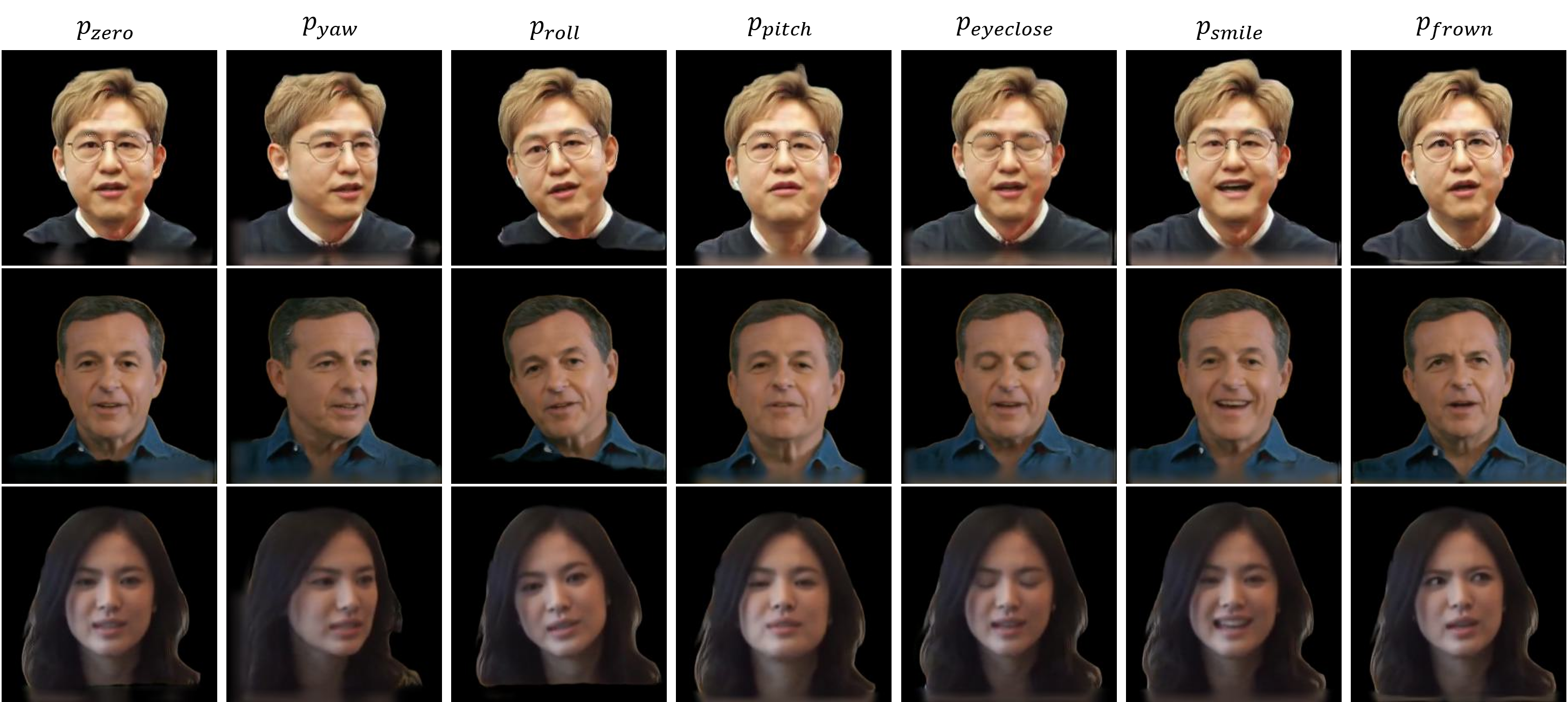}
    \vspace{-0.6cm}
    \caption{Parametric head pose control examples for different target identities, without using driving source images. Different semantic facial expressions and head orientation information are saved in the form of LPMM parameters (first column). These parameters can be applied to different facial identities, editing images in a consistent manner.}
    \vspace{-0.2cm}
    \label{fig:result_params_lpd}
\end{figure*}

\begin{figure*}[t!]
    \centering
    \includegraphics[width=1\linewidth]{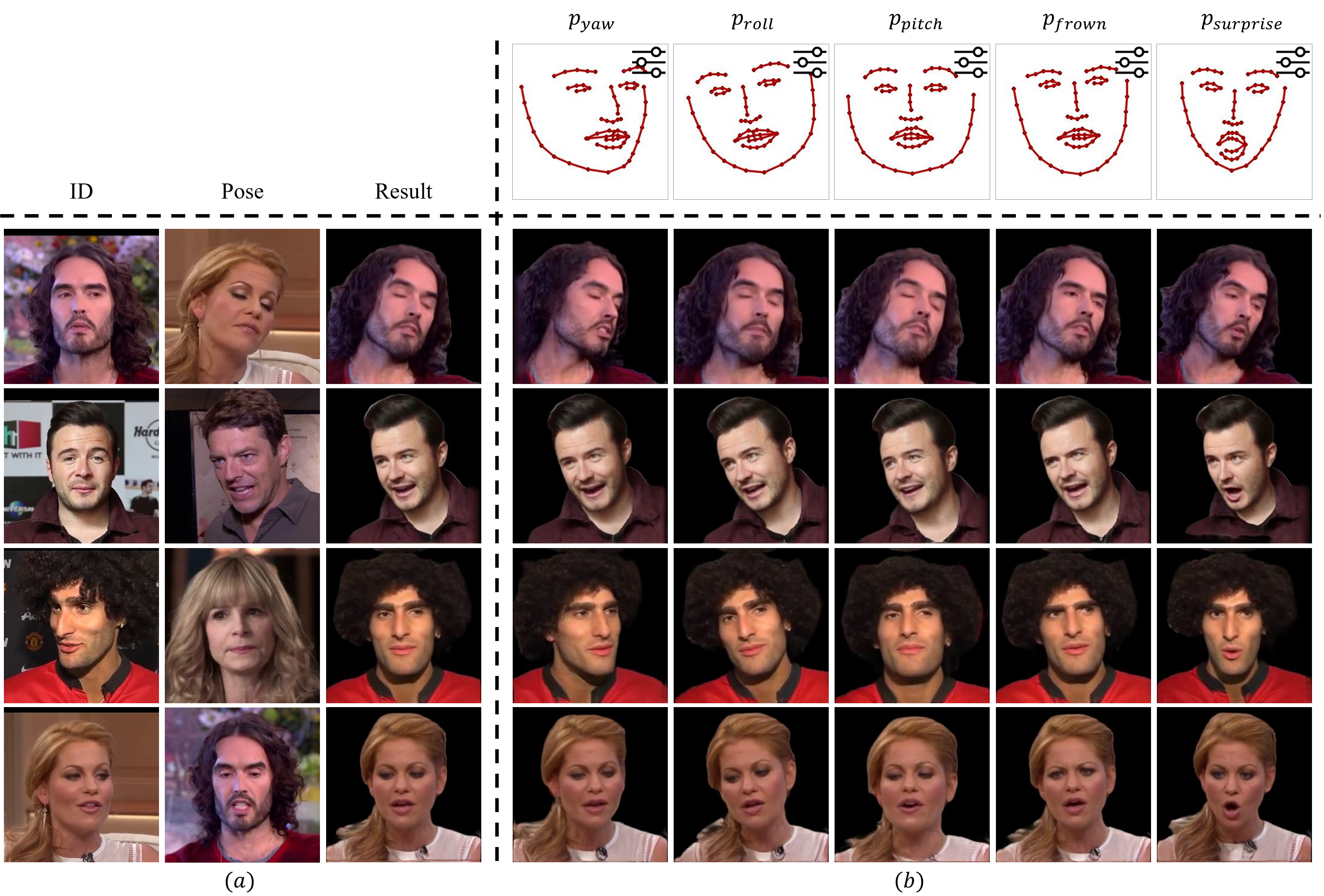}
    \vspace{-0.6cm}
  \caption{Results of image-based head pose edited identities, followed by semantic parameter control. Our method maintains the original inference method of talking head models, which extracts pose information from a driving image as a latent code, and applies it to a facial identity $(a)$. Along with it, pre-defined semantic parameters can be added on to the produced latent code, editing head pose in a reasonable manner while maintaining the original identity $(b)$.}
   \label{fig:drive_param}
\end{figure*}


%% file: section/4_experiment.tex
\section{Experiment}
\label{sec:experiment}



\subsection{Dataset}
\label{subsec:method}
Two talking head video datasets were used for training and evaluation.
We used the VoxCeleb1 dataset~\cite{Nagrani17@vox1} for LPMM and the training pipeline. The dataset consists of YouTube videos, each containing a main speaker identity. For preprocessing, we used the S3FD facial landmark detector ~\cite{zhang2017s3fd} for each video frame to check the perceptibility of a clear face. After collecting around 4.2M images, each image was cropped using the detected bounding boxes, ensuring the center alignment of the face inside the frame. We increased the bounding box size by 80\% and resized to $256 \times 256$. When the increased bounding box size went over the original image boundary, we followed the padding policy of LPD~\cite{burkov2020@lpd}. 

VoxCeleb2 ~\cite{chung2018voxceleb2} contains more identities and videos compared to VoxCeleb1. We collected the videos for 30 identities in VoxCeleb2’s test set, and sampled 64 frames per video, which were then preprocessed following VoxCeleb1’s setting. VoxCeleb2 was used for evaluation, displaying the full potential of our approach.


In addition to VoxCeleb2, we used a separate dataset consisting of Korean celebrity videos and webtoon characters, dubbed as the Korceleb\&Webtoon dataset.

\subsection{Backbone generator}
\label{subsec:generator}





Amongst the possible choices of latent-based neural talking head models that can be used with our method, we chose the two well-known models for our experiments: Latent Pose Descriptors (LPD) ~\cite{burkov2020@lpd} and Latent Image Animator (LIA) ~\cite{wang2021@lia}

\noindent\textbf{LPD.}
LPD uses a combination of identity embedding and pose embedding to generate a talking face.
In practice, the output latent vector of LP-adaptor corresponds to the pose embedding vector $d_p \in \mathbb{R}^{256} $ of the original paper~\cite{burkov2020@lpd}. For LPD, we used a generator fine-tuned with Voxceleb2 or Korceleb\&Webtoon dataset.

\noindent\textbf{LIA.}
LIA interprets motion as an orthogonal set of motion vectors and their corresponding magnitudes.
In practice, the output latent vector of LP-adaptor corresponds to the pose magnitude vector $A_{r \rightarrow d} \in \mathbb{R}^{20}$ of the original paper~\cite{wang2021@lia}.
Since LIA is based on a one-shot setting, a representative identity image was chosen from the dataset as an identity source.



\subsection{Ablation Study}

We evaluate the contributions related to the parameter choices we made in the training of our model.

\noindent\textbf{LP-regressor expressiveness.} For reconstructed landmark accuracy, Normalized Mean Error (NME) was used for measurement. It is defined as
\begin{equation}\label{eq:NME}
    \operatorname{NME}(P, \hat{P})=\frac{1}{N_P} \sum_{i=1}^{N_P} \frac{\left\|p_i-\hat{p}_i\right\|_2}{d},
\end{equation}
where $P$ and $\hat{P}$ denote the predicted and ground-truth coordinates of the respective landmarks, $N_{P}$ is the number of landmarks points, and $d$ is the reference distance to normalize the absolute errors ~\cite{huang202nme}. We use inter-ocular distance as the normalizing factor. We evaluated the NME between the original facial landmark and the reconstructed landmark, for different values of LP-regressor degree $k$. (Table~\ref{Table:k_comparision}) Here, $k=136$ means we used all principal components of the LPMM model. We observed for $k=40$, detailed facial expressions such as eye blinking were well expressed within the landmarks, while maintaining a compact parametric space. For our final solution, we use $k=40$ for its expressiveness.


\subsection{Rig-like control}
\label{subsec:parametric control}

\noindent\textbf{Interactive Parameter Control. } Since LPMM is based on the linear combination of landmark components, each LPMM parameter is independent from one another, and can be manipulated through vector arithmetic. And because LP-adaptor enforces this property with the talking head model, the user can manipulate head pose through linear interpolation. We develop a user-friendly interface where a user can update different LPMM parameter values in the form of interactive sliders, which are fed into LP-adaptor and the neural talking head model to generate a portrait image with the user-defined pose applied, without requiring additional pose-driving images. Figure \ref{fig:result_params_lpd} shows the different results of parametric control through the interface. The results display consistent results for different identity and pose information, proving that our model is capable of being adapted to arbitrary generators.





\input{table/NME_LP-regressor}

\noindent\textbf{Parametric control with additional driving image.}
Since the LP-adaptor’s latent code output $\hat{v}$ can be mixed with the pre-trained talking head model’s pose encoder output $v$, our method retains the property of previous talking-head models to use images as pose driving sources, while applying semantic parametric control at the same time. This can be done by sending the driving image through the LP-regressor and/or the talking-head model’s pose encoder, followed by applying additional specific parameter control. In practice, the user can save pre-defined head pose parameters in the form of blendshapes, and use them directly upon a target facial identity, guaranteeing an efficient, intuitive pose control. 

Figure \ref{fig:teaser} and Figure \ref{fig:drive_param} displays our results with both the driving input image and separate parametric control applied. As shown above, face animations e.g., blinking, surprised, can be reenacted through simple vector arithmetic of predefined parameters, such as $p_{{surprise}}$. This allows users to perform intuitive pose editing without requiring additional driving image sources. In practice, an artist can pre-define different expressional blendshapes for a neutral identity, so that it can be used at any time for different identities.

\noindent\textbf{Base pose visualization.}
To prove that our parametric system is capable of better pose controllability than the latent domain, we visualized the ``base” faces for both domains. We define the parametric ``base” face as the generated face when all parameter values are initialized to zero ($p_{\text{zero}}$), while its latent counterpart will be the generated face of either the mean latent code of the train distribution ($v_{\text{bar}}$), or the zero latent code ($v_{\text{zero}}$).

\begin{figure*}[t!]
    \centering
    \vspace{-0.4cm}
    \includegraphics[width=1\linewidth]{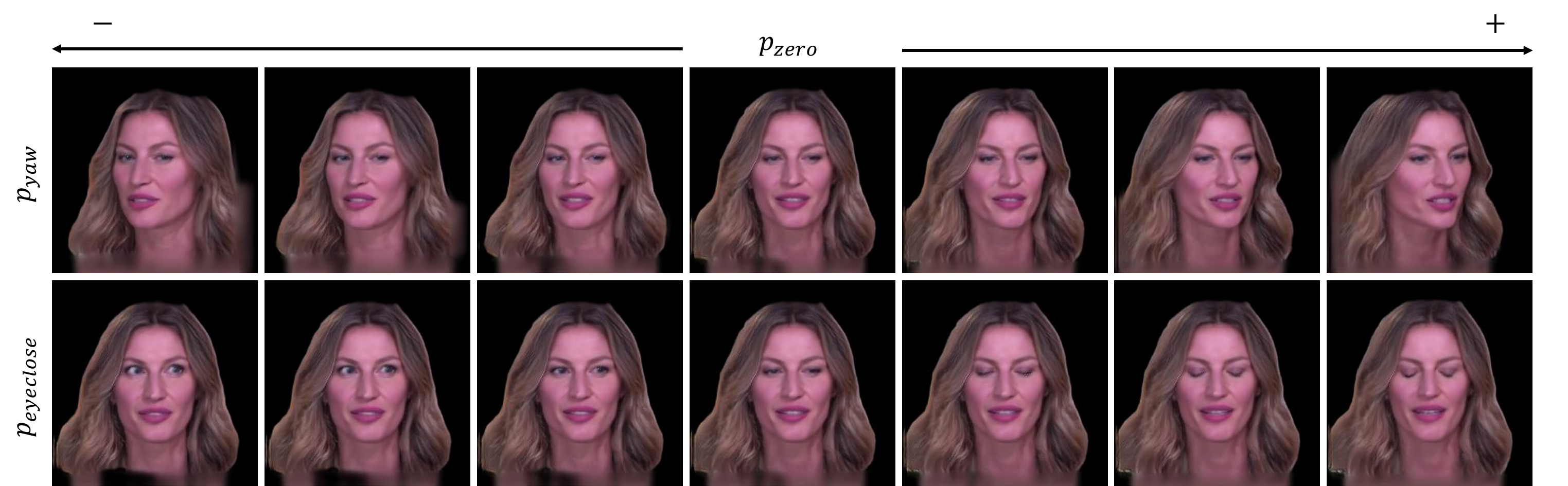}
    \vspace{-0.6cm}
   \caption{Each row shows the parametric interpolation results for head pose. Our system is capable of generating visually smooth and identity preserving interpolation results.}
    \vspace{-0.2cm}
   \label{fig:param_interpolation}
\end{figure*}

\begin{figure}[t]
  \centering
  \includegraphics[width=1\linewidth]{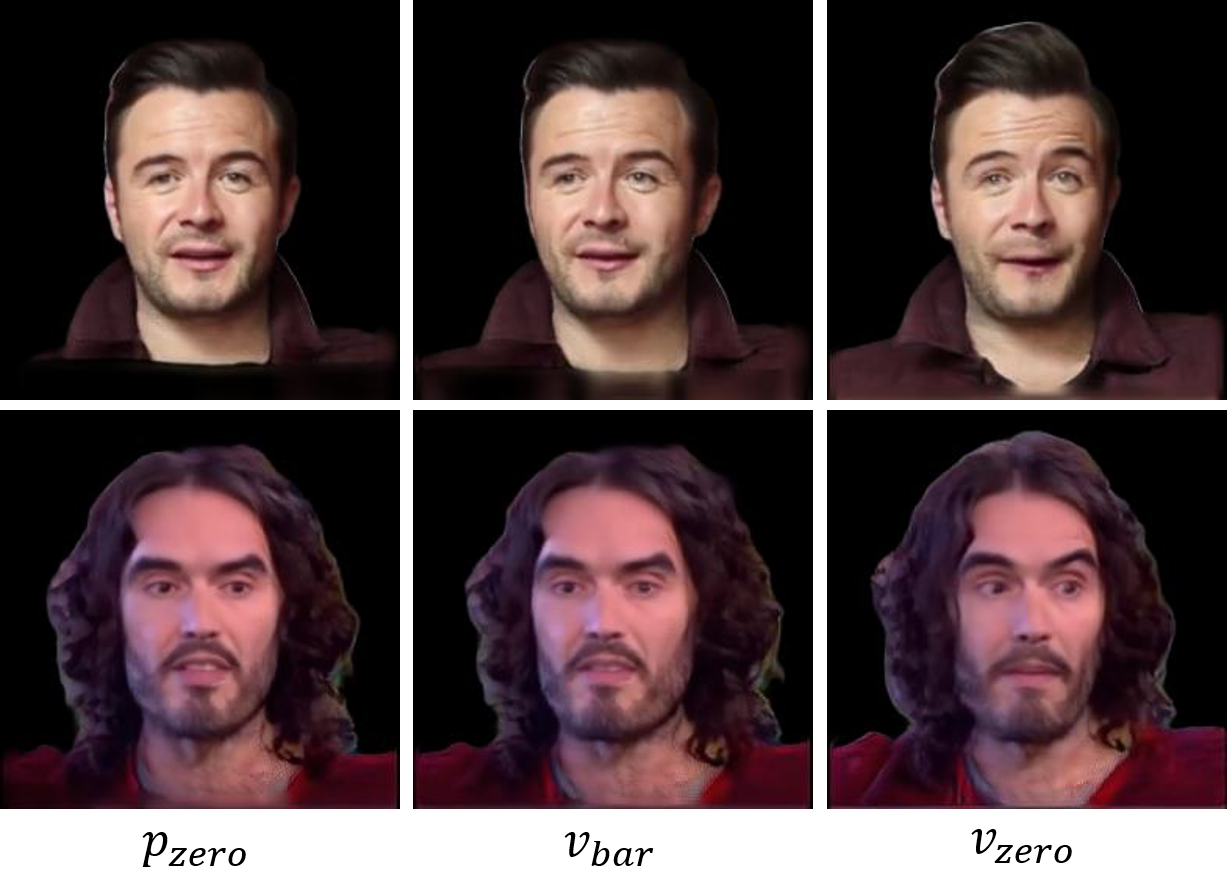}
  \vspace{-0.6cm}
   \caption{Base pose visualization for both parametric space ($p_{\text{zero}}$) and latent space ($v_{\text{zero}}, v_{\text{bar}}$). Compared to $v_{\text{zero}}, v_{\text{bar}}$, the face from $p_{\text{zero}}$ maintains a expression-neutral, frontalized face.}
   \label{fig:base_pose}
\vspace{-0.3cm}
\end{figure}

\begin{figure}
  \centering
  \vspace{-0.6cm}
  \includegraphics[width=1\linewidth]{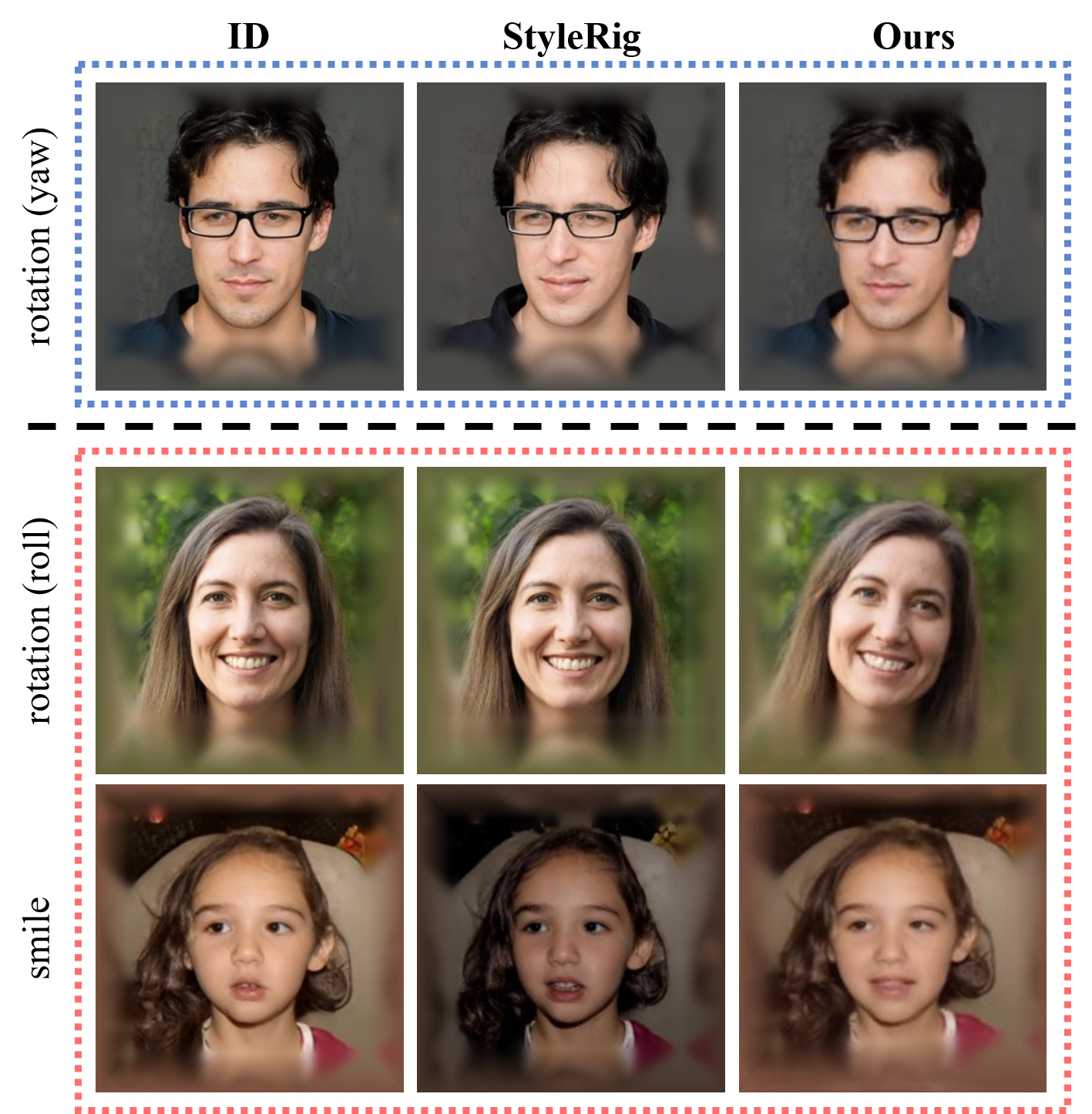}
  \vspace{-0.6cm}
   \caption[]{Comparison to StyleRig ~\cite{tewari2020@stylerig}. The StyleRig results from the first row was created from our StyleRig implementations, while the two bottom row results were collected from the original paper. Our approach is on par with StyleRig for head orientation (blue), and produces a better pose transfer for in-place rotations and complex expressions (red).}
   \label{fig:stylerig_comparison}
  \vspace{-0.3cm}
\end{figure}

Figure \ref{fig:base_pose} visualizes the ``base” face for different facial identities. It can be observed that while both $v_{\text{bar}}$ and $v_{\text{zero}}$ are biased towards a certain head pose and expression, $p_{\text{zero}}$ maintains a neutral, face-frontalized face, displaying that the parametric domain has a consistent starting point for pose manipulation. The existence of a robust base face also allows smooth pose interpolation to be done by a simple scalar multiplication to the pose parameters (Figure \ref{fig:param_interpolation}), unlike previous latent manipulation approaches ~\cite{burkov2020@lpd} which requires two pose vectors to perform the interpolation.

\subsection{Comparison with previous methods}
\label{subsec:stylerig}



Since there is no publicly-available code base or checkpoints available for StyleRig, we trained and implemented StyleRig by ourselves, following the practices mentioned in the original paper. While implementing, we noticed that, unlike our method's pipeline, StyleRig's pipeline is based on two different data distributions; 3D facial scanning data for 3DMM ~\cite{blanz1999@3dmm}, and photorealistic portrait images ~\cite{karras2019stylegan} for StyleGAN, which creates a discrepancy between the parametric and latent space domain.
Figure \ref{fig:stylerig_comparison} shows the comparison between StyleRig parametric control results and our results.
We use pose edited results both from the original paper\footnote[1]{The StyleRig results for the third row of Figure \ref{fig:stylerig_comparison} shows a different level of illumination. This is because the original paper also meant to capture illumination along with head pose.} and from our StyleRig implementation. It can be observed that while StyleRig succeeds in editing the yaw angles of the image (first row), it fails to control the roll movements (second row), which was attributed to a bias introduced in StyleRig's training data ~\cite{tewari2020@stylerig}.
Our method does not suffer from such biases and is capable of precise control over the head orientation.
For expression editing, StyleRig often leads to incorrect expression mapping for certain expressions (third row). Compared to StyleRig, our model produces better editability towards portrait images, especially for extreme facial expressions.


%% file: table/NME_LP-regressor.tex
\begin{table}[]
    \centering
    \resizebox{\columnwidth}{!}{%
    \begin{tabular}{lccccc}
    \multicolumn{6}{c}{\textbf{LP-regressor NME}}                                                   \\ \hline
    \multicolumn{1}{l|}{\textbf{Eval dataset}} & k=5  & k=10 & k=20 & k=40          & k=136         \\ \hline
    \multicolumn{1}{l|}{Voxceleb2}             & 4.33 & 2.89 & 2.38 & 2.29          & \textbf{2.24} \\ \hline
    \multicolumn{1}{l|}{Korceleb\&Webtoon}      & 5.81 & 3.92 & 3.18 & \textbf{3.13} & 3.25         
    \end{tabular}%
    }
    \caption{For different values of LP-regressor degree $k$, we evaluated the NME between the original facial landmark and the reconstructed landmark.}
    \label{Table:k_comparision}
\end{table}

%% file: section/5_conclusion.tex
\section{Discussion}
\label{sec:conclusion}


We have presented a novel method to provide semantic head pose control over fixed neural talking head model. Unlike the previous approach on semantic pose control  ~\cite{tewari2020@stylerig}, our method utilizes talking head model as a back-bone generator. Since, it can bring the pose-identity disentanglement power and does not require complex training pipeline. One limitation of our method is that to enable semantic control over facial expression, we might have to discover a combination of parameters to manipulate an intuitive expression, instead of controlling a single parameter value. We note, however, that head orientations can be frontalized, and for real-world applications our method does not require users to find such parameter values.